\documentclass[letterpaper,twocolumn,10pt]{article}
\usepackage{usenix2019_v3}

\usepackage{graphicx}
\usepackage{booktabs}
\usepackage{amsmath}
\usepackage{amssymb}
\usepackage{algorithm}
\usepackage{algorithmic}

\providecommand{\citep}{\cite}
\providecommand{\citet}{\cite}

\begin{document}

\date{}

\title{\Large \bf Measuring the Wrong Thing: Internal Harmfulness Scores
Anti-Rank Successful Jailbreaks}

\author{
{\rm Mingyu Luo$^{1,\dagger}$\quad
Ming Deng$^{1,2,\dagger}$\quad
Zilang Qiu$^{1,3}$\quad
Yiming Cheng$^{4}$\quad
Ci Tao$^{1}$\quad
Xue Tan$^{1}$}\\
{\rm Sijin Sun$^{5}$\quad
Yangfu Li$^{6}$\quad
Ping Chen$^{7,*}$\quad
Jun Dai$^{8}$\quad
Xiaoyan Sun$^{8}$}\\[0.8ex]
{\rm\small $^{1}$College of Computer Science and Artificial Intelligence, Fudan University}\\
{\rm\small $^{2}$School of Computer Engineering and Science, Shanghai University\quad
$^{3}$Beijing Normal University}\\
{\rm\small $^{4}$Tsinghua University\quad
$^{5}$Institute of Advanced Intelligence and Computing, A*STAR}\\
{\rm\small $^{6}$School of Communication and Electronic Engineering, East China Normal University}\\
{\rm\small $^{7}$Institute of Big Data, Fudan University\quad
$^{8}$Department of Computer Science, Worcester Polytechnic Institute}\\[0.4ex]
{\rm\small $^{\dagger}$These authors contributed equally.\quad
$^{*}$Corresponding author.}
}

\maketitle

\begin{abstract}
Internal safety scores judge a prompt before any text is generated, and they
are validated by how well they separate harmful prompts from benign ones. That
separation is then read as evidence that the score will also catch the attacks
that succeed. Harmful intent is a property of the prompt. Jailbreak success is
an outcome produced later by a particular target model, decoding policy, and
judge. A filter tuned on a score that measures the wrong quantity spends its
false positive budget on attacks that would have failed anyway. In this paper
we audit that inference. Attention based measurements are usually read from
prompt dependent locations, so a wrapper changes both the content being judged
and the place the signal is taken from. We therefore introduce Active Attention
Probing, which supplies a fixed content independent measurement coordinate. We
pair every base goal with a plain and a wrapped version and generate real
completions from the target models. On Llama, wrapping raises harmful
generation from $0.05$ to $0.27$ while harmful intent AUROC falls from $0.936$
to $0.803$, so the attacks grow more dangerous while the prompts look safer to
the score. Among wrapped harmful prompts the outcome AUROC is $0.220$, which
places the attacks that succeeded below the attacks that failed. Rare token,
passive, and detector derived channels reproduce the reversal on the same
matched design, and the reversal itself persists across three target models,
seven attack families, and two independent judges. Distribution
shift then degrades calibration and threshold transfer before it degrades
ranking.
\end{abstract}

\section{Introduction}

An internal safety score can identify harmful intent before generation. A
deployed filter may then treat a high score as evidence that the prompt will
successfully jailbreak its target. These interpretations refer to different
quantities. Harmful intent is a property of the prompt. Realized success also
depends on the target model, decoding policy, and judge.

Our Llama results show that the two quantities can diverge. Wrapping a harmful
request raises the harmful generation rate from $0.05$ to $0.27$. At the same
time, harmful intent AUROC falls from $0.936$ to $0.803$. Among wrapped harmful
prompts, the score places successful attacks below failed attacks and reaches an
outcome AUROC of $0.220$. These results do not show that prompt safety detectors
fail at their stated task. They show that harmfulness validation does not
establish validity for realized success.

We test this inference with matched plain and wrapped versions of each base
goal. We generate real completions from each target model and label the realized
outcome. Attention based scores are usually read from prompt dependent
locations, so a wrapper changes both the content being evaluated and the place
the signal is taken from. A score change is then difficult to attribute to the
model rather than to the moved coordinate. Active Attention Probing (AAP)
removes this confound by providing a fixed content independent coordinate in
the system region, so a prompt and its wrapped version are compared at a common
reference point. AAP is the methodological contribution that makes the audit
possible. The scientific claim of the paper concerns what the resulting score
predicts. Rare token, passive, system span attention, and refusal logit
channels test whether that claim depends on the AAP coordinate.

\begin{figure*}[t]
\centering
\includegraphics[width=0.97\textwidth]{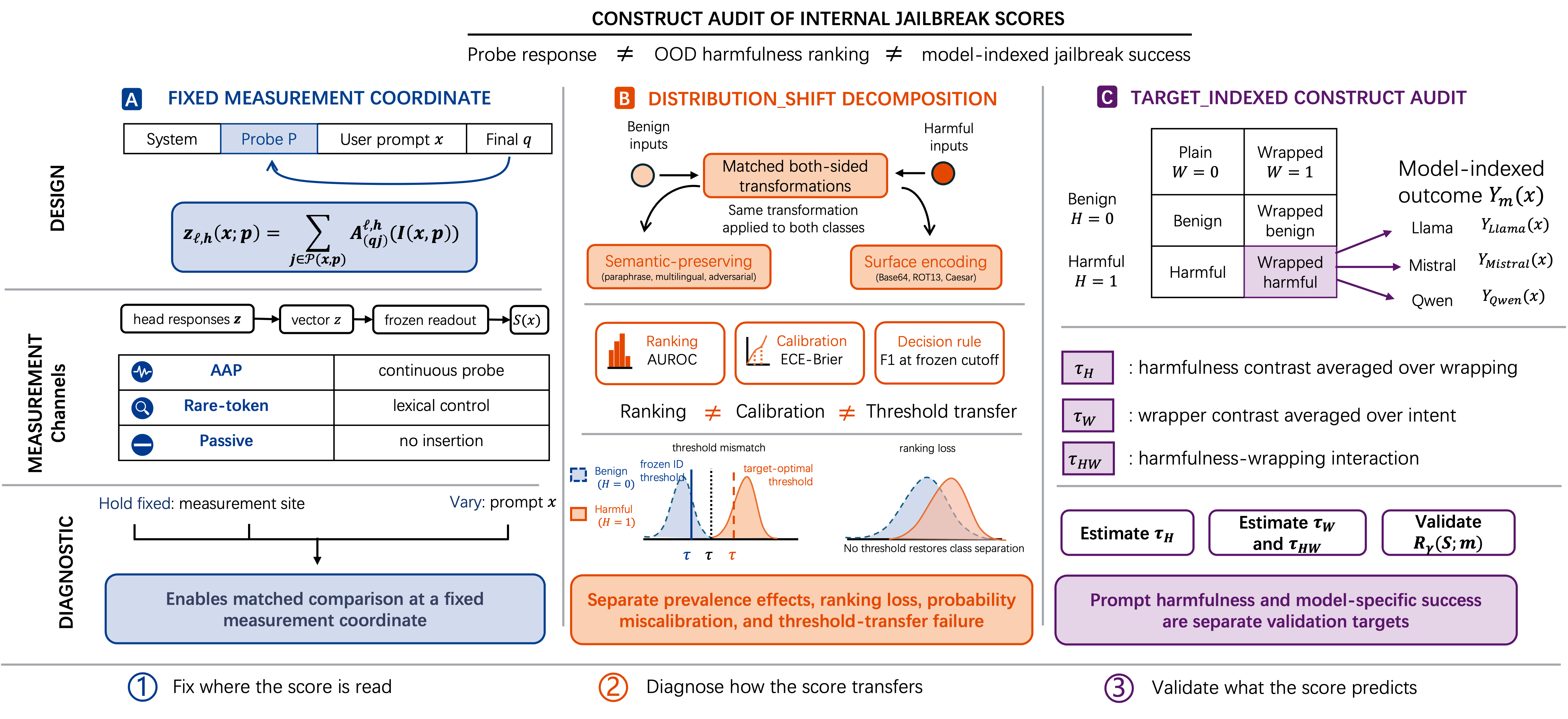}
\caption{The audit reads one score against three questions. Each column is a
question, and its rows give the design, the measurement channel, and the
diagnostic. Panel A asks whether the response is measured at a stable place, so
AAP fixes the coordinate and rare token and passive channels match it. Panel B
asks whether the harmful intent reading survives a change of distribution, so
every transformation is applied to both classes and ranking, calibration, and
threshold transfer are separated. Panel C asks whether the score anticipates
what the target does, so it crosses $H$ with $W$ over matched goals and
validates against target indexed outcomes $Y_m$. Panel C is where the two
constructs come apart.}
\label{fig:audit-overview}
\end{figure*}

Figure~\ref{fig:audit-overview} presents the audit. RQ1 asks whether a score
validated on harmful intent can rank target specific jailbreak success. RQ2
then asks whether the original harmfulness interpretation survives distribution
shift. We evaluate ranking, calibration, and threshold transfer separately
because each can fail while the others remain useful.

Prior work shows that effective jailbreaks can suppress internal harmfulness
features \citep{ball2024understanding,lin2024representation}. Probes trained on
success labels can also transfer poorly across attack families
\citep{kirch2025features}. We ask what a score trained on harmfulness can
validly support. We make three contributions.

\begin{itemize}
    \item We distinguish prompt harmfulness from target specific jailbreak
    success and estimate both quantities on matched prompts with realized
    outcomes.
    \item We show that a strong harmfulness score can rank realized success in
    the opposite direction. The result persists across measurement channels,
    targets, attacks, and independent judges. We also quantify the effect on a
    deployed filter.
    \item We test whether the original harmfulness interpretation survives
    distribution shift. The audit separates ranking loss, probability
    miscalibration, and threshold transfer.
\end{itemize}

\section{Related Work}

\noindent\textbf{Internal safety scores.} Attention Tracker and
AttentionDefense use attention to detect prompt injection or adversarial prompts
\citep{hung2025attentiontracker,sun2024attentiondefense}. HiddenDetect,
GradSafe, refusal directions, mutation sensitivity, and deployed filters use
logits, gradients, hidden states, prompt variants, or layered classifiers
\citep{jiang2024hiddendetect,xie2024gradsafe,arditi2024refusal,
robustharmful2025,zhang2024jailguard,rebuff2023}. This line of work establishes
that internal signals discriminate prompt level safety or attack labels. We ask
the separate question of whether such a score is a valid predictor of a
realized, target indexed jailbreak outcome. We audit the interpretation
supported by the existing validation and do not dispute performance on the
stated prompt classification task.

Our audit treats each detector as a representation, a readout, and a threshold.
A prompt level label validates harmful intent for any of these channels. It
cannot establish whether the target model will comply. GradSafe comes closest
and compares gradients against a hypothetical compliant response that the target
model never generates \citep{xie2024gradsafe}. HiddenDetect reads a logit lens
over multimodal inputs \citep{jiang2024hiddendetect}, and our HiddenDetect style
row ports that readout to text only prompts with every deviation listed in the
appendix. Attention head specialization under attack has been studied directly
\citep{robustharmful2025}. Our per head decomposition makes no specialization
claim.

\noindent\textbf{Jailbreak outcomes.} Effective jailbreaks can suppress
internal harmfulness representations and move harmful prompts toward a harmless
region \citep{ball2024understanding,lin2024representation}. Prompt side probes
trained directly on success labels can transfer below chance across attack
families \citep{kirch2025features}. That result uses Llama-3.1-8B, which is also
our primary target. Our positive control asks a narrower question. It
tests whether success information is available within template based wrappers
over a shared goal set. It does not claim transfer across attack paradigms. Our
main audit asks whether a score trained on harmfulness already has the
orientation and operating meaning required for success prediction.

Jailbreak benchmarks report attack success for one generator and one judge. We retain both indices instead of treating success as a prompt property.

\noindent\textbf{Validity under distribution shift.} Safety detectors can
degrade as attack styles and source distributions change
\citep{lee2018mahalanobis,jailbreaksovertime2025,sahoo2026external,
lin2024benchmarkslie}. Single source splits also overstate safety detector AUC
by $8.0$ to $16.5$ points \citep{lin2024benchmarkslie}. That work proposes leave
one dataset out evaluation. The leave one dataset out control removes corpus
identity. We apply each transformation to both classes so that the
transformation cannot serve as the label. Our both sided design removes
transformation identity. No transformation experiment in that work addresses
this shortcut. We also separate ranking,
calibration, and threshold transfer. This decomposition identifies which part
of the score interpretation fails.

AUROC is invariant to monotone score transformations while calibration and fixed thresholds are not. One fixed threshold metric therefore cannot support all three claims. That failure can be silent. Safety classifiers on frozen embeddings fall from $85\%$ to chance ROC-AUC while mean confidence falls only $14\%$ \citep{sahoo2026external}.

\noindent\textbf{Probing as measurement.} AAP uses a continuous embedding in
the soft prompt and prefix tuning tradition
\citep{lester2021power,li2021prefix}. It provides a fixed measurement anchor
instead of steering model behavior \citep{zou2023representation}. Passive,
lexical, and norm matched controls test whether insertion, lexical content, or
learning explains the result \citep{hewitt2019designing}. Response magnitude
alone does not establish construct validity, so we validate every channel on
held out labels. Every realized outcome is generated without the probe.

\section{Measurement Setting and Estimands}\label{sec:threat}

\noindent\textbf{Setting.} We study a pre generation internal score computed
from one eager attention prefill pass. The score has access to attention and
hidden states. It sees the prompt $x$ but not the realized completion.

\noindent\textbf{Measurement targets.} Let $H(x)\in\{0,1\}$ denote harmful
intent in the base request, $W(x)\in\{0,1\}$ the presence of a jailbreak
wrapper, and $Y_m(x)\in\{0,1\}$ whether target model $m$ actually emits a
harmful completion under a frozen safety judge. We omit judge and decoding
indices unless we vary them. An ineffective jailbreak has
$W{=}H{=}1,Y_m{=}0$. An effective jailbreak has $W{=}H{=}Y_m{=}1$. Evidence
that a score measures $H$ establishes construct validity for harmful intent.
Prediction of $Y_m$ requires separate evidence. RQ1 audits that criterion
claim. RQ2 tests whether validity for $H$ survives a change in input
distribution.

\noindent\textbf{Validity criteria.} Construct validity asks whether the score ranks its training label. Outcome validity asks whether it ranks $Y_m$ for a named target and judge. Decision validity asks whether a threshold keeps its operating behavior. Evidence for an earlier claim does not imply a later one.

\noindent\textbf{Deployment scope.} We study a prompt side filter that blocks when the score exceeds a threshold. The relevant error includes successful attacks that score below failed attacks. A fixed benign false positive budget measures how the ranking allocates blocking capacity between the two groups.

\section{Audit Design}\label{sec:method}

Passive attention detectors aggregate locations determined by the prompt
\citep{hung2025attentiontracker,sun2024attentiondefense}. A wrapper can
therefore change both the input and the measurement location. AAP fixes this
location. Let $x=(x_1,\ldots,x_T)$ denote a tokenized chat prompt and let
$p_\theta\in\mathbb R^{K\times d}$ denote a short probe embedding.
$I(x,p_\theta)$ inserts the probe into the system region after the chat template
is materialized. The user tokens remain unchanged. AAP reads the attention
response at the probe and maps it to a scalar score. A matched passive read uses
the same template boundary without inserting the probe.

\noindent\textbf{Attention response.} For layer $\ell$, head $h$, probe
positions $P(x,p)$, and final input position $q$, the response is
\begin{equation}
z_{\ell,h}(x,p)
= \sum_{j\in P(x,p)} A^{\ell,h}_{qj}\!\left(I(x,p)\right).
\label{eq:probe-response}
\end{equation}
The vector $\mathbf z(x,p)$ concatenates the monitored heads. XGBoost maps this
vector to a score $S(x)\in[0,1]$. Rare token, common token, norm matched, system
span, passive attention, and refusal logit channels test changes in probe
content, insertion, location, and feature family. Response magnitude appears in
the appendix.

\noindent\textbf{Training and separation.} The primary probe has $K{=}3$ rows.
We train it only on the BeaverTails training split and freeze it before every
audit. No JailbreakHub \citep{shen2024doanythingnow}, WildJailbreak, or
JailbreakBench prompt selects the
probe, heads, or harmfulness readout. Every derived shift set starts from the
held out test split. The harmfulness readout uses fixed constants and no
validation search. The outcome supervised positive control selects
regularization inside each cross validation fold. The appendix gives the
optimization objective, initialization, head sensitivity, and continuous
controls.

The probe starts from safety semantic token embeddings. The objective maximizes the safe versus harmful attention gap on the training split and uses only the training selected $20$ head set. The dense RQ2 read and every held out domain stay unseen.

Three controls separate prompt surface, response magnitude, and transferred
ranking. Head selection is the first. Sparse head subsets are high variance
across random draws, so RQ2 uses the dense read. Prompt surface is the second. A
length only classifier gives AUROC from $0.49$ to $0.62$, including settings
where encoding makes prompts five times longer, so length does not carry the
signal. Correlation between derived sets is the third. Source clustered
intervals are $0.94$ times as wide as unclustered intervals, so shared source
prompts do not inflate significance.

Scores use \texttt{inputs\_embeds} with the probe present. All generations use
the original prompt and ordinary chat template without the probe. No reported
outcome is produced by a probe perturbed generation. The appendix measures
the insertion effect. Next token symmetric KL has median $0.22$. Refusal flips occur on $7.5\%$ of prompts. A
separate generation audit estimates a change in attack success of $-0.03$ with
a $95\%$ CI of $[-0.070,0.007]$.

The appendix gives each control its read location, feature family, and readout. The safety semantic initialization alone does not reproduce the learned response. The gain comes from optimization and not from safety semantic content. These
controls change one measurement choice at a time. They do not reimplement the
published systems that motivated them.

\noindent\textbf{Matched target design.} We cross harmful intent and wrapping
while holding the base goal fixed. For $h,w\in\{0,1\}$, define
\begin{equation}
\mu_{hw}=\mathbb E[S(x)\mid H(x)=h,W(x)=w].
\label{eq:target-cells}
\end{equation}
The balanced effects are
\begin{equation}
\begin{aligned}
\tau_H&=\tfrac12[(\mu_{10}-\mu_{00})+(\mu_{11}-\mu_{01})],\\
\tau_W&=\tfrac12[(\mu_{01}-\mu_{00})+(\mu_{11}-\mu_{10})],\\
\tau_{HW}&=\mu_{11}-\mu_{10}-\mu_{01}+\mu_{00}.
\end{aligned}
\label{eq:target-effects}
\end{equation}
$\tau_H$ measures harmfulness, $\tau_W$ measures the paired wrapper effect, and
$\tau_{HW}$ measures whether wrapping changes the harmfulness response.
$R_H$ denotes harmful intent AUROC. $R_Y(S,m)$ denotes outcome AUROC for
effective and ineffective harmful prompts under target $m$. The same design
applies to Llama, Mistral, Qwen, Gemma, and Phi.

The attention grids differ in layer and head count across these models. Equation~\ref{eq:probe-response} stays unchanged. Fixed score target
substitution changes the generator while holding prompts, score source,
decoding, and judge fixed. The target native matrix also changes the score
source. This distinction separates generator effects from score architecture
effects.

RQ1 uses the $20$ heads selected on the training split. RQ2 uses the dense set of
all heads. Both specifications read the same extracted attention tensor and both produce an outcome AUROC below chance. A benign
quantile rule fixes the nominal false positive rate for the deployment audit
\citep{angelopoulos2021conformal,massart1990tight}. The appendix provides the
finite sample bound, subset sensitivity, and the check that the deployed score
recomputes from that tensor.

The rule uses the empirical $1-\alpha$ score quantile of $m$ safe calibration prompts. Sampling error and benign shift both loosen the guarantee. The audit therefore measures upper tail stability. It
does not provide evidence about outcome validity.

\section{Evaluation}\label{sec:eval}

\noindent\textbf{Setup.} Llama-3.1-8B-Instruct is the primary target. Mistral,
Qwen, Gemma, and Phi test transfer across attention architectures. We curate
BeaverTails \citep{ji2023beavertails} by removing duplicates and conflicting
labels. The resulting pool contains $1774$ safe and $3671$ harmful prompts. A
balanced set of $1000$ prompts per class forms the training split. The held out
split contains $774$ safe and $2671$ harmful prompts. We split source prompts
before creating paraphrase, multilingual, or encoding variants.

\noindent\textbf{Metrics.} AUROC measures ranking. Brier score, ECE, and
logistic recalibration slope measure probability calibration. Fixed and
refitted F1 measure threshold transfer. Paired AUROC differences use DeLong
tests and bootstrap $95\%$ intervals. AUROC counts tied scores at one half. An
ideal probability has a recalibration slope of one and an intercept of zero. We
also report the F1 gap between a threshold frozen on the source domain and an
oracle target threshold. The gap diagnoses the operating rule. It does not
change the ranking score. The appendix reports every prespecified cell, the
resampling units, and the complete statistical protocol.

\subsection{Measurement Validation}\label{sec:indist}

Before auditing what the score predicts, we establish that AAP is a meaningful
and controlled measurement channel. This subsection validates the framework both
audit questions rely on and makes no claim about what the score predicts.
The frozen harmfulness readout reaches a train to test AUROC of $0.985$.
Cross validation within the held out pool gives $0.957$ to $0.982$ across the
five architectures. We treat the latter range as a separability check because
each fold fits part of the held out pool. A paired intervention study also
shows that learned probes increase response magnitude over lexical and norm
matched controls without consistently improving ranking. We therefore treat
AAP, rare token, passive, system span, and refusal logit reads as parallel
measurement channels. The audit does not use response amplification as evidence
of validity.

Norm matched controls confirm this distinction. Learned probes raise the response by at least sixteen times over a matched random probe on all three architectures, while linear readout AUROC moves by at most $0.06$ and favors the random probe on Llama. Norm matching therefore removes embedding norm as the sole explanation for amplification. A larger response does not guarantee a better ranking.

\begin{figure}[t]
\centering
\includegraphics[width=1\columnwidth]{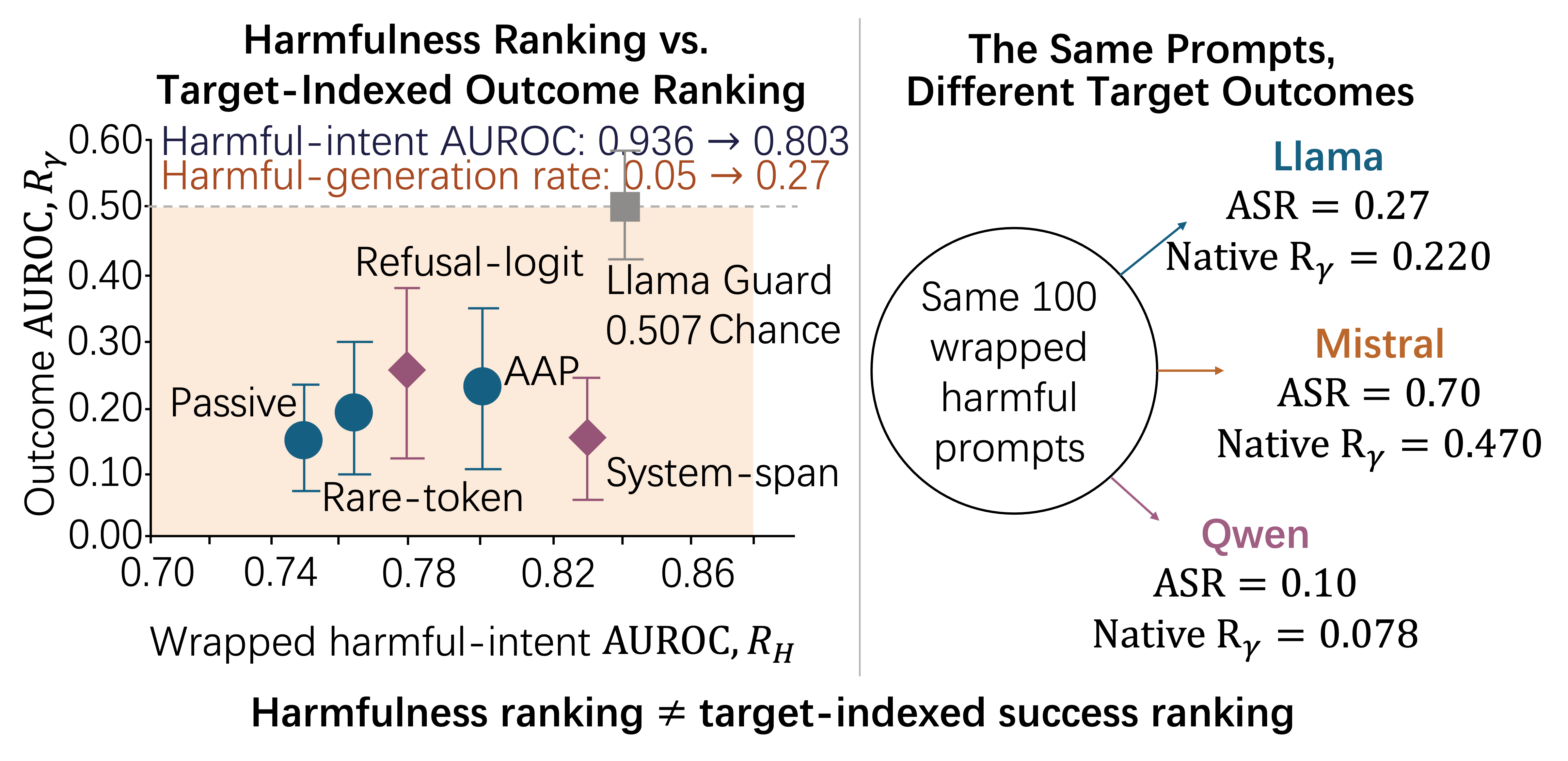}
\caption{Harmful intent ranking and realized success ranking separate. The left
panel shows harmful intent AUROC $R_H$ and target outcome AUROC $R_Y$ for each
channel. Intervals are bootstrap $95\%$ CIs and the dashed line marks chance.
The right panel shows target specific success rates and outcome rankings for
the same $100$ wrapped harmful prompts.}
\label{fig:construct-outcome}
\end{figure}

\subsection{RQ1: Does Harmfulness Validity Extend to Realized Success?}
\label{sec:target-decomp}

We pair the $100$ benign and $100$ harmful goals in JailbreakBench
\citep{chao2024jailbreakbench} with plain and wrapped versions. Five wrapper
families are balanced across intent classes. We freeze prompts, assignments,
success criteria, and greedy decoding before inference. Llama Guard 3 provides
the primary outcome labels \citep{inan2023llamaguard,grattafiori2024llama3}. A
frozen Qwen3-14B rubric relabels the cached completions.

\noindent\textbf{The score measures harmful intent but reverses success.}
Figure~\ref{fig:construct-outcome} states the finding in one picture, and the
paragraphs below give the estimates behind it.
AAP has a harmfulness effect of $\tau_H=0.521$ with CI $[0.458,0.584]$, a
wrapper effect of $\tau_W=-0.094$, and an interaction of
$\tau_{HW}=-0.257$ with CI $[-0.359,-0.158]$. Wrapping selectively lowers the
score for harmful requests. Harmful intent AUROC falls from $0.936$ to $0.803$
while Llama harmful generation rises from $0.05$ to $0.27$. Among the $100$
wrapped harmful goals, the score ranks the $27$ successful attacks below the
$73$ failures. Outcome AUROC is $0.220$ with CI $[0.109,0.343]$.

The induced outcome analysis reaches the same conclusion. Among $95$ goals that
Llama refuses without a wrapper, $23$ become successful after wrapping. A McNemar test on the paired plain and wrapped outcomes gives $p=3\times10^{-6}$. The
score ranks these induced successes below the remaining failures with
$R_Y=0.254$ and CI $[0.129,0.396]$. The plain harmful prompts also show an
inverse ranking with $R_Y=0.128$ and CI $[0.025,0.259]$. The wrapper therefore
does not create the reversal. What carries the signal is the wrapped score
level rather than the plain to wrapped change, since ranking induced success by
that paired difference gives an AUROC of only $0.421$. A goal level logistic model keeps the coefficient negative after it
controls the plain score, plain outcome, and wrapper family. Coefficients range
from $-0.59$ to $-1.33$ across Llama and Qwen specifications. Within family centering gives $R_Y=0.333$ with CI $[0.236,0.444]$. Family stratified pooling, family fixed effects, and leave one family out estimates all agree. Successful attacks have the lower mean score in all five families.
Three families contain only two successes, so we do not estimate a separate
AUROC for each family. Five sampled decodes give $R_Y$ from $0.202$ to $0.252$ under three outcome definitions. Every interval excludes chance and the greedy label is representative.
The appendix reports the full induced, family, and decoding analyses.

\noindent\textbf{The mismatch changes filter behavior.} We set the threshold
using benign prompts at a nominal $5\%$ false positive rate. The achieved rate
is $5.7\%$. The filter removes $42.5\%$ of failed attacks but only $14.8\%$ of
successful attacks. The gap is $-27.7$ points with CI $[-44.8,-8.8]$. The
filter admits $23$ of the $27$ realized jailbreaks. On the strong attack cohort,
$92\%$ of successful attacks pass under either retained judge. Absolute harm
still falls from $27$ to $23$, and the score removes about $2.6$ times as many
successful attacks as random blocking for the same benign budget. The score is
not worse than no filter. It spends the budget mainly on attacks that already
fail. The harmful rate among admitted prompts rises from $0.270$ to $0.354$.
At a nominal $10\%$ false positive rate the achieved rate is $11.1\%$. The
filter blocks $80.8\%$ of failed attacks and $29.6\%$ of successful ones. The
gap is $-51.2$ points. Preexisting misses explain $53\%$ of leakage at the
$5\%$ point. The wrapper main effect explains $23\%$, and the inverse
ranking explains $24\%$. The inverse share rises to $53\%$ at the $10\%$ point.
An outcome supervised score reduces the admitted harm rate to $0.062$ at
matched coverage. Rare token and passive channels show the same allocation
pattern.

\noindent\textbf{Success information is available before generation.} An
outcome supervised readout uses the same internal features and grouped cross
validation. Twenty five repeats keep every base goal within one fold. The
readout reaches out of fold AUROC of $0.930$ on wrapped harmful prompts and
$0.939$ on all harmful prompts. It reaches $0.875$ with CI $[0.800,0.939]$ when
the test wrapper family is held out and $0.725$ with CI $[0.680,0.788]$ on
WildJailbreak. Nested cross validation reproduces these values. A label
permutation null gives $0.497\pm0.072$. The best
factorial baseline receives wrapper family success rates and still trails the
readout by $0.156$ on wrapped prompts. On WildJailbreak, a word level TF-IDF
logistic model is the strongest baseline at $0.651$ under the same protocol and
regularization grid. The internal readout exceeds it by a paired margin from
$0.016$ to $0.139$. These results show that the internal coordinates contain
target specific success information. Qwen3 relabeling lowers the positive
control ceiling to $0.883$ but leaves the harmfulness score outcome AUROC at
$0.220$. Judge dependence limits the positive control. Qwen3 identifies $14$ successful
prompts, all within the $27$ prompts identified by Llama Guard. Their agreement
has $\kappa=0.61$. The harmfulness score keeps the same outcome AUROC under both
judges, but the Qwen3 labels cannot distinguish within family success from the
family success rate. We therefore use the positive control to establish the
presence of outcome information. We use Llama Guard alone for the stronger
claim about the goal level component.

\noindent\textbf{The result extends beyond AAP.} Rare token, passive, system
span, and refusal logit channels retain wrapped harmfulness AUROC from $0.751$
to $0.828$. Their outcome AUROC ranges from $0.145$ to $0.248$. Each channel also has a negative harmfulness by wrapper interaction and an outcome interval entirely below chance. The passive channel inserts no tokens, so probe insertion does not create either effect. Llama Guard
remains a strong harmfulness discriminator and gives an outcome AUROC of
$0.507$. On $350$ natural WildJailbreak prompts with $133$ effective attacks,
AAP outcome AUROC is $0.359$. Llama Guard reaches $0.625$ on the same cohort.
That value exceeds every internal channel and its own matched design value.
Rare token, passive, and perplexity channels give $0.374$, $0.449$, and
$0.487$ \citep{alon2023detecting}. The matched design remains the basis for the main claim.
Across five probe seeds, wrapped harmfulness AUROC stays from $0.770$ to
$0.861$ and outcome AUROC stays from $0.130$ to $0.246$. Per head analysis localizes the mismatch with Spearman $\rho=-0.85$ across $1024$ heads. The appendix reports the mechanical null that limits this analysis to a descriptive role.

\noindent\textbf{Targets change the outcome, not the conclusion.}
Table~\ref{tab:native-matrix} recomputes the score on Llama, Mistral, and Qwen.
Six of nine intervals lie below chance and none lies above chance. Mistral already complies with $0.68$ of plain harmful requests, so wrapping barely raises its success rate. Its outcome
AUROC near chance is therefore uninformative about an induced attack effect.
Fourteen Mistral goals change from failure to success and twelve change in the
opposite direction. McNemar testing gives $p=0.85$. Five sampled decodes also
show that target outcomes disagree at a rate of $0.81$ against a no target
effect null of $0.29$. Pairwise native score correlations are only $0.42$ to
$0.54$. Native recomputation does not recover a positive outcome ranking.

\begin{table}[tb]
\centering
\footnotesize
\setlength{\tabcolsep}{1mm}
\begin{tabular}{@{}lccc@{}}
\toprule
Score & Llama & Mistral & Qwen \\
\midrule
Llama   & \shortstack{.220\\{[.109,.343]}} & \shortstack{.491\\{[.363,.623]}} & \shortstack{.163\\{[.024,.358]}} \\
Mistral & \shortstack{.292\\{[.183,.411]}} & \shortstack{.470\\{[.345,.598]}} & \shortstack{.190\\{[.061,.343]}} \\
Qwen    & \shortstack{.172\\{[.085,.275]}} & \shortstack{.386\\{[.272,.505]}} & \shortstack{.078\\{[.027,.139]}} \\
\midrule
Wrapped success rate & .27 & .70 & .10 \\
\bottomrule
\end{tabular}
\caption{Outcome AUROC $R_Y$ for the same $100$ wrapped harmful goals. Rows
identify the native score and columns identify the generator. Entries include
bootstrap $95\%$ CIs.}
\label{tab:native-matrix}
\end{table}

\noindent\textbf{Strong attacks and independent judges reproduce the pattern.}
We evaluate GCG, AutoDAN, AutoDAN-Turbo, PAIR, TAP, Best-of-N, and Many-shot
\citep{zou2023universal,liu2024autodan,liu2025autodanturbo,chao2023pair,
mehrotra2024tap,hughes2024bestofn,anil2024manyshot}. Table~\ref{tab:strong-attacks}
reports Llama Guard 3, Qwen3, and a StrongREJECT rubric
\citep{souly2024strongreject}. We exclude the
StrongREJECT column from inference because the available substitute marks many
verbatim refusals as successful. That leaves fourteen attack cells, which with
the nine target cells of Table~\ref{tab:native-matrix} make twenty three in
total. Six of the nine and six of the fourteen have intervals below chance, and
no interval lies above chance. A bootstrap clustered by goal and attack family
gives a pooled outcome AUROC of $0.313$ with CI $[0.212,0.422]$ and
$p=0.0006$. Those twenty three cells share one pool of $200$ base goals, and the
same clustered bootstrap gives a design effect of $9.3$, or about $2.5$
independent units, so we report no sign test over cells. Qwen3 is independent of
the
search procedure for GCG, AutoDAN, PAIR, and TAP. GCG and AutoDAN remain below
chance under Qwen3 after BH FDR correction.
Pooled logistic models also control the plain goal score and attack family.
They retain negative score coefficients of $-0.54$ under Llama Guard and
$-0.35$ under Qwen3. Base goal clustered intervals exclude zero for both
judges.

Every attack row pairs the plain and attacked version of the same goal under the frozen Llama scorer. The $n=200$ rows combine JailbreakBench and HarmBench \citep{mazeika2024harmbench}. The StrongREJECT substitute marks about $30\%$ of verbatim refusals as successful and also fires on benign controls. The inverse ranking survives strict consensus and union labels in the appendix. We found a checkpoint defect in our own pipeline and regenerated five search based rows. No conclusion changes. The appendix reports the repair and every interval.

Qwen3 agrees with Llama Guard on $0.87$ of wrapped Llama outcomes and on only $0.66$ for Mistral. It preserves the AAP, rare token, and passive inverse rankings on Llama. These
results show that one judge does not create the matched reversal. They do not
remove the broader uncertainty from automated outcome labels.

\begin{table}[tb]
\centering
\footnotesize
\setlength{\tabcolsep}{4.0pt}
\begin{tabular}{@{}lrccc@{}}
\toprule
\multicolumn{5}{@{}l}{\textbf{A. Attack success rate}} \\
\addlinespace[1pt]
Attack & $n$ & LG3 & SR & Q3 \\
\midrule
Best-of-N      & 200 & .10$\to$.94 & .30$\to$.73 & .04$\to$.26 \\
Many-shot      & 200 & .10$\to$.18 & .32$\to$.30 & .03$\to$.18 \\
GCG            & 100 & .11$\to$.17 & .28$\to$.25 & .02$\to$.09 \\
AutoDAN        & 100 & .05$\to$.34 & .29$\to$.49 & .03$\to$.20 \\
PAIR           & 100 & .05$\to$.52 & .26$\to$.78 & .01$\to$.07 \\
TAP            &  50 & .02$\to$.16 & .20$\to$.46 & .00$\to$.06 \\
AutoDAN-Turbo  &  50 & .02$\to$.38 & .22$\to$.51 & .00$\to$.14 \\
\midrule
\multicolumn{5}{@{}l}{\textbf{B. Outcome AUROC $R_Y$}} \\
\addlinespace[1pt]
Attack & $n$ & LG3 & SR & Q3 \\
\midrule
Best-of-N      & 200 & .35 & .43 & .54 \\
Many-shot      & 200 & .48 & .53 & .49 \\
GCG            & 100 & .19 & .27 & .26 \\
AutoDAN        & 100 & .20 & .14 & .24 \\
PAIR           & 100 & .48 & .38 & .43 \\
TAP            &  50 & .31 & .25 & .21 \\
AutoDAN-Turbo  &  50 & .28 & .25 & .51 \\
\bottomrule
\end{tabular}
\caption{Strong attack results on Llama. Panel A reports plain to attacked
success rates. Panel B reports outcome AUROC. LG3 and Q3 are Llama Guard 3 and
Qwen3. SR is shown for transparency and excluded from inference.}
\label{tab:strong-attacks}
\end{table}

\noindent\textbf{Attention dilution does not explain the reversal.}
Fixing the probe position does not fix its share of the softmax, and attacks
can target that share directly \citep{wang2024attngcg}.
Wrapping lowers total attention to the system region from $0.383$ to $0.225$.
Dividing by system region attention is a clean dilution control because that
region holds a constant $13$ tokens and the wrapper lands in the user turn, so
the denominator is a fixed competition set. The outcome ranking remains inverse
after that normalization and after normalizing by the full attention row.
Sequence length is not the driver either, since prompt length explains only
$4\%$ of score variance. The appendix reports both normalizations, the
retrained normalized readout, and the probe perturbation audit.

Together, these results are consistent with a representational decoupling
account. Wrapping may weaken features used by a readout trained on harmful
intent while separately changing the model processes that govern compliance.
The passive and refusal logit results argue against probe insertion as the
source, and the outcome supervised readout shows that success information
remains available. We treat this account as a hypothesis because the audit does
not identify a unique causal pathway.

\noindent\fbox{\parbox{0.955\columnwidth}{\small\textbf{RQ1 finding.} A score
that ranks harmful intent well can rank realized jailbreak success in the
opposite direction. Harmfulness validation does not establish validity for
success prediction.}}

\subsection{RQ2: Does Harmfulness Validity Survive Distribution Shift?}
\label{sec:g1}

RQ1 shows that validating a score against one construct does not license
reading it as a predictor of another. We now ask whether the score stays valid
for the construct it does measure once the input distribution changes. A shift
can break that reading in three ways. The ranking of harmful above benign can
fail, the numerical meaning can fail while the ranking holds, or only the
transfer of a previously chosen threshold can fail while both hold. A single
fixed threshold number cannot say which occurred.

We therefore keep the construct fixed as harmful intent and vary the input
distribution. Meaning preserving shifts include WildJailbreak, paraphrases,
and English, Chinese, French, and Japanese translations
\citep{jiang2024wildteaming}. Surface transformations include base64, rot13,
and caesar encoding. We apply every transformation to both classes. This design
prevents the transformation from serving as the label.

Readout family is a material source of variation. Logistic regression and a
linear SVM match the source AUROC near $0.98$. Their mean AUROC across the same
shifted domains falls to $0.55$ and $0.64$, compared with $0.75$ for XGBoost.
On the Japanese condition, the linear readouts reach $0.01$ to $0.08$ and
XGBoost reaches $0.228$. We keep XGBoost fixed as the primary readout and give
the analysis in the appendix.

The Japanese result is an inversion with a known transformation artifact.
Translated prompts have $2.4$ times the character trigram similarity of the English source text. The Japanese transformation therefore collapses prompts
toward a shared register. Its orientation free AUROC is $1-0.228=0.772$, close
to the Chinese and French range. We retain the row as a readout sensitivity
case and do not treat it as clean evidence of semantic robustness failure.

\begin{table}[tb]
\centering
\footnotesize
\setlength{\tabcolsep}{2.2pt}
\begin{tabular}{@{}lccccc@{}}
\toprule
Domain & AUROC & ECE & slope & F1@0.5 & F1@$\tau_{20\%}$ \\
\midrule
In distribution      & 0.985 & 0.045 & 0.54 & 0.925 & 0.937 \\
WildJailbreak         & 0.850 & 0.366 & 0.51 & 0.559 & 0.664 \\
Paraphrase            & 0.858 & 0.203 & 0.32 & 0.734 & 0.769 \\
English round trip   & 0.987 & 0.041 & 0.76 & 0.950 & 0.950 \\
Multilingual (zh)     & 0.802 & 0.247 & 0.43 & 0.731 & 0.719 \\
Multilingual (fr)     & 0.778 & 0.232 & 0.42 & 0.688 & 0.749 \\
\bottomrule
\end{tabular}
\caption{Ranking, calibration, and threshold transfer under meaning preserving
shifts. F1@0.5 freezes the source threshold. F1@$\tau_{20\%}$ relocates only
the threshold using a labeled target slice.}
\label{tab:calib}
\end{table}

\noindent\textbf{Thresholds fail before rankings.}
Table~\ref{tab:calib} shows that the frozen decision rule loses more F1 than the
score loses AUROC on meaning preserving shifts. The source readout is not a
calibrated probability. Its slope is $0.545$. The slope column is an in sample Cox fit. Platt scaling on a frozen source
slice changes the slope to $1.019$ and ECE from $0.045$ to $0.009$. The
calibrated source rule reaches WildJailbreak F1 of $0.667$ without target
labels. The oracle reaches $0.678$. The gap remains from $0.006$ to $0.020$ on
domains where AUROC stays from $0.76$ to $0.86$. This repair is not universal.
On the English round trip control, source calibration changes ECE from $0.041$
to $0.171$ and slope from $0.761$ to $1.488$. The frozen calibration map can
therefore harm a domain that already ranks well. The appendix reports the
remaining calibration checks.

The appendix separates a score that was never calibrated from a score whose calibration changes under shift. The source map repairs the source
domain and improves five shifted domains. It does not restore a probability
interpretation everywhere. The encoding slopes remain near zero because the
ranking has disappeared. The Japanese slope remains negative because the class
ordering has reversed. No monotone calibration map can repair either case. The ranking loss also replicates across architectures. Mistral, Qwen, Gemma, and Phi repeat the WildJailbreak evaluation and their AUROC falls to a range of $0.60$ to $0.85$.

\noindent\textbf{Prevalence explains only part of threshold shift.}
We reweight each domain to a common balanced class prior. AUROC is unchanged.
The source common prior threshold reaches WildJailbreak F1 of $0.790$, compared
with an oracle value of $0.796$. HiddenDetect and refusal direction scores keep
larger threshold gaps under the same audit, while the AttentionDefense style
channel is better calibrated than AAP on WildJailbreak. Every score family
loses separation on both class base64. A benign only quantile rule holds its
nominal false positive rate in distribution and nearly matches oracle recall on
paraphrase and English prompts. Under benign shift its WildJailbreak recall
falls to $0.03$ against an oracle of $0.25$. This identifies benign
tail stability as a separate requirement. The appendix reports the per score
common prior table.

\noindent\textbf{Both class encoding removes harmfulness separation.}
Encoding only harmful prompts gives AUROC from $0.85$ to $0.89$. A model free character entropy statistic reaches AUROC $0.997$ to $1.000$ on that single class comparison and $0.513$ on unencoded prompts. Encoding both
classes gives $0.51$, $0.50$, and $0.52$ for base64, rot13, and caesar. Mistral,
Qwen, Gemma, and Phi reproduce a mean AUROC from $0.50$ to $0.51$. The
single class design measures a transformation cue. The both class design shows
that the score loses harmfulness separation when the cue is unavailable.

\noindent\fbox{\parbox{0.955\columnwidth}{\small\textbf{RQ2 finding.}
Distribution shift can change calibration and threshold transfer while ranking
survives. Stronger surface shifts can also remove the harmfulness ranking
itself.}}

\section{Discussion and Limitations}\label{sec:discussion}

\noindent\textbf{Implications.} A benchmark should name the construct attached
to each label: harmfulness detection requires $H$ and $R_H$, whereas realized
success requires generations and $Y_m$ indexed by target, decoding policy, and
judge. A complete audit should report harmful-intent ranking on held-out
prompts, realized outcomes for protected targets, and blocking at a fixed
benign budget. The evaluated thresholds reduce absolute harm but allocate
blocks poorly within the harmful cohort, assigning lower scores to successful
attacks; low harmfulness therefore does not imply refusal.

\noindent\textbf{Limitations.} Outcome labels are automated, without a blinded
human audit, and three-judge agreement is low (Fleiss
$\kappa\approx0.3$); judge choice thus changes $Y_m$ but does not itself support
our thesis. Llama Guard guides several attack searches, leaving only the rubric
judges independent for those rows. Primary outcomes use one greedy decode; five
sampled decodes preserve the factorial result, whereas the strong-attack and
shift results remain single-draw. The Llama and Qwen cells contain only $27$
and $10$ positives, and the targets are not difficulty-matched, so they show a
range of behavior rather than a controlled comparison. Strong attacks and
derived channels are evaluated on Llama; only Best-of-N and Many-shot were
repeated on Mistral and Qwen. Probe insertion perturbs the prefill, although
outcomes use unmodified prompts. The Japanese transform contains an artifact,
and the lowest OOD results depend on readout family.

\noindent\textbf{Scope and artifacts.} The estimates cover only the tested
prompts and checkpoints, so new deployments require new outcome audits. The
matched factorial design supports the main claim. WildJailbreak and strong
attacks are uncontrolled external checks. We release an artifact
containing the analysis code, configurations, tests, and publication-aligned
numeric aggregates at
\url{https://github.com/kuangren777/aap-audit}.

\section{Conclusion}

A strong harmful intent AUROC does not establish that a score can predict
jailbreak success. Our matched audit shows that the score can place successful
attacks below failed attacks. The result persists across measurement channels,
targets, attack families, and independent judges. Distribution shift can also
change the original harmfulness interpretation. Harmful intent and realized
success require separate validation.

\bibliographystyle{plain}
\bibliography{aaai2027}

\appendix
\section{Additional Ablations}\label{sec:ablate}
These ablations support the measurement and evaluation choices in the main paper.

\noindent\textbf{Probe optimization.} The learned $K{=}3$ continuous probe is
initialized from safety-semantic token embeddings and optimized only on the
BeaverTails training split with Adam, learning rate $10^{-3}$, $50$ optimizer
updates, each on a freshly sampled mini-batch of $20$ prompts per class. The objective is
\begin{equation}
\begin{aligned}
\mathcal L(p)&=-\frac{1}{|\mathcal H|K}\sum_{h,k}
\left|\bar z^{\mathrm{safe}}_{h,k}(p)-
\bar z^{\mathrm{attack}}_{h,k}(p)\right|\\
&\quad+0.1\,\bar z^{\mathrm{safe}}(p)
+0.01\,\|p\|_{\mathrm{row}}.
\end{aligned}
\end{equation}
The first term maximizes the absolute source-domain attention gap, the second
discourages high safe-prompt attention, and the last controls row-wise embedding
norm. No OOD or target-audit prompt is used for probe or readout selection.

\noindent\textbf{Benign-only quantile rule.} For trusted safe calibration prompts
$\mathcal C_0=\{x_i\}_{i=1}^m$ and false-positive budget $\alpha$, we set
\begin{equation}
\begin{aligned}
\widehat\tau_\alpha
&=\widehat Q_{1-\alpha}
\bigl(\{s_\theta(x_i;p):x_i\in\mathcal C_0\}\bigr),\\
\widehat y(x)
&=\mathbb I[s_\theta(x;p)>\widehat\tau_\alpha].
\end{aligned}
\end{equation}
The Dvoretzky--Kiefer--Wolfowitz bound gives sampling
error $\sqrt{\log(2/\delta)/(2m)}$. Benign shift adds
$\gamma_d=\sup_t|F_{0,d}(t)-F_{0,\mathrm{cal}}(t)|$. The diagnostic therefore
identifies upper-tail stability rather than outcome validity.

\noindent\textbf{Head-coverage sensitivity.} Sparse $20$-head subsets are high-variance. Across $40$ random $20$-head draws the minimum AUROC over nine OOD domains is $0.60\pm0.15$ (95\% CI $[0.25,0.83]$), and the layer-diverse and safety-ranked heuristics fall at the $65$th and $47$th percentiles of that distribution, inside the noise. We therefore report the all-head read as a fixed measurement specification and treat subset choice as a sensitivity analysis.

\noindent\textbf{Readout sensitivity.} The in-distribution result is similar for XGBoost, logistic regression, and linear SVM (5-fold AUROC ${\approx}0.98$). Under shift, however, the dense all-head result depends on the readout family: linear probes on the full feature degrade far more than the tree readout on the lowest-ranked both-sided domains (Japanese: $0.01$--$0.08$ against XGBoost's $0.228$). We therefore report XGBoost as the fixed primary readout and make this dependence explicit.

\noindent\textbf{Surface-form and correlation controls.} Prompt length does not drive the signal: a length-only classifier is near chance everywhere ($0.49$--$0.62$ AUROC), including on encoding where attacks are $5\times$ longer. We recompute confidence intervals with a bootstrap clustered by source prompt, because the encoding, paraphrase and multilingual sets share source prompts. The clustered interval is $0.94\times$ as wide as the unclustered interval (a $6\%$ narrowing), so within-source correlation does not inflate our significance. Finally, exploratory retraining shows that increasing the in-distribution attention gap can reduce OOD AUROC toward chance. Response magnitude and transferred ranking are therefore empirically distinct.

\noindent\textbf{Norm-matched continuous controls.} We construct a random
three-vector probe with each row normalized to the corresponding learned row
norm. It therefore has the same length, insertion coordinate, dtype, and
per-row embedding norm as the learned probe. On $100+100$ BeaverTails prompts,
the Llama learned/random controls give response magnitude $1.008/0.046$ and
linear-readout AUROC $0.931/0.944$. Mistral gives $2.054/0.124$ and
$0.964/0.906$. Qwen gives $0.802/0.030$ and $0.932/0.896$. The control rules
out embedding norm as the sole source of amplification, while the differing
ranking outcomes illustrate the architecture-dependent amplification--ranking
interaction. Qwen's non-finite entries affect fewer than $0.2\%$ of responses
and are excluded only from aggregate estimation.

\noindent\textbf{Relative-anchor feature.} We also tested a relative feature
$r_{\ell,h}=\log\!\big((A^{\rm probe}_{\ell,h}+\epsilon)/(A^{\rm ctrl}_{\ell,h}+\epsilon)\big)$
that normalizes probe attention by the matched natural-control read at the same
head, intended to cancel sequence-length and global-attention drift across domains.
On WildJailbreak (with per-prompt aligned probe/control features), the relative
feature alone \emph{underperforms} the raw probe read (AUROC $0.744$ vs $0.803$),
while concatenating raw and relative features gives only a marginal gain ($0.816$).
The relative feature varies across domains and the concatenated representation
changes WildJailbreak AUROC only marginally. The reported analysis therefore
uses the pre-specified raw attention read.

\noindent\textbf{Factorial intervention details.} The nine paired conditions are a natural
three-token system control. Rare probes with $1$, $3$, and $5$ IDs. A shuffled
rare probe. Independently sampled rare IDs. Common IDs. Safety-semantic IDs,
and the default rare triplet inserted at the user midpoint. Linear probes on
the full head-response vector remain strong in distribution (mean five-fold
AUROC $0.95$--$0.99$ across the three displayed conditions and models), even
when the scalar mean response has the opposite orientation. This confirms that
the information is multivariate and cautions against interpreting the sign of
an all-head average as classifier performance. Qwen produced non-finite values
for fewer than $0.11\%$ of head responses in these runs. Raw arrays were
retained, while only summary estimation replaced non-finite entries by zero.
Substituting exclusion or median imputation for the zero fill changes the
reported statistics by less than their quoted precision, since the affected
fraction is under $0.2\%$.

\noindent\textbf{Hybrid representation.} A hybrid variant concatenates the
final-position projection onto a refusal direction and the hidden-state norm to
the attention response. It is not used for any RQ1 or RQ2 estimate in the main
paper. We report it only because the in-distribution measurement-validation
numbers quote both variants, where it tracks the attention-only readout to
within $0.005$ AUROC on every architecture.

\noindent\textbf{Strong-attack budgets and goal subsets.} The seven attacks do
not share one goal list, which is why their unattacked success rates differ. The
$n=200$ rows (Best-of-N, Many-shot) pool JailbreakBench and HarmBench harmful
goals. GCG, AutoDAN and PAIR use $100$ goals each and TAP and AutoDAN-Turbo use
$50$, drawn from the same pool but not identical subsets, so the plain-prompt
rate in Panel A varies by row (for example $0.11$ for GCG against $0.05$ for
AutoDAN under Llama Guard) and rows should be compared to their own plain
column rather than to one another. Search budgets were fixed before running:
GCG uses $40$ optimization rounds with early stopping once the target negative
log-likelihood falls below $0.10$. The tree and lifelong-memory searches (TAP,
AutoDAN-Turbo) run at depth, width and branching factor $2$ because of their
cost. Best-of-N and Many-shot use their published sampling and shot budgets. All
greedy attacks decode $256$ new tokens, while query and sampling attacks retain
their search-time winning completion.

\noindent\textbf{Goal-difficulty control.} The pooled model quoted in the main
text is, for each judge,
\begin{equation}
\operatorname{logit}\Pr(Y_i=1)=\beta_0+\beta_S\widetilde S_i+
\beta_P\widetilde S_i^{\rm plain}+\gamma_{a(i)},
\end{equation}
where $\widetilde S_i$ is the standardized score of the attacked prompt,
$\widetilde S^{\rm plain}_i$ the standardized score of the same base goal
unattacked, and $\gamma_{a}$ an attack-family effect. Intervals come from a
bootstrap that resamples base goals as clusters.

\section{A Checkpoint Defect and Its Repair}

While auditing the strong-attack table we found, and then fixed, a real defect
in our own pipeline. We report it in full because it changes five rows of that
table and because the failure mode is easy to reproduce in any resumable
evaluation harness.

\noindent\textbf{Mechanism.} Both the scoring and the generation stages resumed
from a checkpoint that was validated by \emph{length only}, restarting at
\texttt{len(cached)} in a prompt list rebuilt from a mutable manifest. Each
attack script also has a smoke-test truncation. A pilot run over $s$ goals
produced a short manifest and complete checkpoints. The subsequent full run
overwrote the manifest with $B$ goals and silently reused the stale checkpoints
as a prefix, displacing records by exactly $k=2(B-s)$ positions. The displaced block lands on the wrong goals, because the records are laid out
with the benign class first.

\noindent\textbf{Proof and scope.} Per-prompt scores are computed one prompt at
a time on a frozen detector, so records sharing a base goal must be byte-equal
across cells. Under the fitted $k$ the displaced value equals the consensus
value at position $i+k$ for every disagreeing record, on all five stored
artifacts and in all five affected cells, and the fitted $k$ matches $2(B-s)$
in each. Five of twelve cells are affected, namely GCG, AutoDAN, AutoDAN-Turbo, PAIR
and TAP, all on the Llama target. Best-of-N and Many-shot are clean on all three
targets. The attack stage itself is immune because its checkpoint is keyed by
goal identifier, so no attacked prompt needed to be recomputed. The defect
touched $240$ completions and $320$ scores. The score stage is corrupted for
more records than the completion stage because scoring runs over all records
while generation runs only over those still missing a completion.

\noindent\textbf{Effect on the results.} We regenerated the affected cells with
an assignment bound to the record identifier, preserving each attack's original
completion semantics: GCG and AutoDAN are greedy throughout, whereas the query
attacks retain their search-time winning completion, whose identity binding we
re-verified against the attack checkpoint ($400/400$ matched on both prompt and
completion). Ten of the fifteen repaired attack-by-judge rows are bit-identical to the
published ones, because most of the displaced records fell in the benign half of
the record list, which neither ASR nor $R_Y$ reads. The five that move are the
three GCG rows and the rubric entries for AutoDAN and AutoDAN-Turbo. GCG moves and
strengthens: $R_Y$ becomes $0.189$/$0.264$ under Llama Guard and Qwen3, and its
Qwen3 interval now excludes chance where it previously did not. Among the fourteen retained attack-by-judge cells the count significantly below
chance is unchanged at six, with GCG--Qwen3 entering and TAP--Llama Guard
leaving. No conclusion in the paper changes.

\section{Per-Head Decomposition and Its Mechanical Null}

\noindent\textbf{Per-head decomposition.} We re-extracted the per-head probe
response for the locked $400$-prompt factorial cohort as a dense tensor of shape
$400\times32\times32\times3$ (prompt $\times$ layer $\times$ head $\times$ probe
position). Recomputing the deployed scalar score from this tensor reproduces the
cached per-prompt scores bitwise (Pearson $r=1.000$, maximum absolute deviation
$0$ over all $400$ prompts), so the per-head features come from exactly the
computation that produced every reported number. Three independent extractions
were byte-identical.

Scoring each of the $1024$ heads on its own, per-head harmful-intent AUROC
and per-head outcome AUROC vary widely, and the two are strongly
anti-correlated (Pearson $-0.845$, Spearman $-0.849$). The five heads that rank
intent best average $R_Y=0.117$ and the twenty best average $R_Y=0.121$. We do
not quote any individual head's value: with $27$ positives, a maximum taken over
$1024$ heads is not interpretable in sample.

This anti-correlation is mostly mechanical, and we say so rather than reading it
as functional specialization. Any head that ranks intent well would anti-rank
outcome with no specialization involved at all, were the outcome a deterministic
threshold on the aggregate score. Simulating exactly that null, matched to the
observed positive count and to the observed aggregate $R_Y$, already produces a
mean head-level Spearman correlation of $-0.71$ (95\% range $[-0.83,-0.48]$)
along the published score direction, and $-0.82$ along an alternative aggregate
direction, where the null range covers our observed value outright. Our
observed $-0.849$ sits outside the null band for the published direction
($p=0.006$) and the partial correlation controlling each head's loading on the
aggregate direction is $-0.59$, also outside its null band, so a residual
component survives, but the effect is not robust to the choice of aggregate
direction, and we therefore report the decomposition as a localization of where
the two criteria diverge rather than as evidence that distinct heads encode
distinct constructs. By layer band, early heads (layers $0$--$10$) are at chance
on both criteria ($R_H$ $0.510$, $R_Y$ $0.510$), while middle ($11$--$21$) and
late ($22$--$31$) heads discriminate intent and anti-rank outcome
($0.708/0.267$ and $0.724/0.260$ as head means, $0.098$ and $0.104$ when the
band is aggregated into a single score). Outcome-informative heads nevertheless
exist: $266$ heads have $R_Y>0.5$ in sample and $22$ survive a max-statistic
permutation correction at the family-wise level, against roughly five expected
under the matched null. The dense all-head read
gives $R_H=0.882$ with $R_Y=0.087$, and the deployed twenty-head read gives
$0.803$ with $0.220$. The qualitative conclusion is the same under both.

\noindent\textbf{Wrapper-family stratification.} Three of the five wrapper
families produce only two successes each, so no per-family outcome AUROC is
estimable and the marginal estimate is dominated by families with more events.
Within-family centering gives $R_Y=0.333$ (CI $[0.236,0.444]$, bootstrap
$p=0.003$, permutation $p=0.011$) and Mantel--Haenszel stratified pooling gives
$0.323$ (CI $[0.178,0.472]$, $p=0.020$). Leave-one-family-out estimates range
over $0.296$--$0.345$. Only $15.4\%$ of the $1971$ discordant pairs are
within-family, which is why the stratified estimate is the less precise of the
two. Restricting to the two adequately powered families gives $0.313$ with an
interval that just touches chance ($[0.144,0.503]$). Between families, mean
score and attack success are negatively related but not significantly so with
five points (Pearson $-0.77$, $p=0.13$).

We deliberately do not rest homogeneity on a non-significant interaction test.
A power simulation calibrated to the observed family intercepts shows that the
score-by-family interaction likelihood-ratio test on these data has a rejection
rate of ${\approx}0.16$ both under a true null and under an alternative in which
one family's slope is fully sign-flipped: its power equals its type-I rate, so a
non-rejection carries no information. The homogeneity evidence we do report is
sign consistency: in all five families the successful attacks have the lower
mean score (differences $-0.34$, $-0.36$, $-0.31$, $-0.10$, $-0.03$).

\noindent\textbf{Outcome-supervised readout.} Training on the outcome label with
grouped repeated cross-validation (25 repeats, base goal never spanning folds)
gives out-of-fold AUROC $0.930$ on the wrapped harmful cohort, $0.939$ on all
harmful prompts, and $0.725$ (CI $[0.680,0.788]$) on the 350-prompt
WildJailbreak audit, against label-permutation nulls of $0.50\pm0.07$
($p=0.005$ in each case). Leave-one-wrapper-family-out gives $0.875$
(CI $[0.800,0.939]$). Nested cross-validation reproduces these values, so
selecting the regularization strength is not the source of the advantage. We
quote bootstrap intervals rather than the spread across cross-validation
repeats, which is not a confidence interval because the folds are not
independent.

Every baseline is refitted under the readout's own protocol and its own
regularization grid, and the two cohorts must be read separately. On the
JailbreakBench factorial cohorts the baselines span $0.62$--$0.80$. The
strongest is not a text model but an oracle that is told each wrapper family's
empirical success rate, and the readout still leads it by $+0.156$ (wrapped
cohort) and $+0.134$ (all harmful). On WildJailbreak neither a wrapper-identity
nor a plain-goal baseline exists, since those prompts are naturally occurring
adversarial phrasings with no synthetic family and no matched unwrapped
counterpart. There the baselines reach only $0.48$--$0.65$, the best being a
word-level TF-IDF logistic model at $0.651$, a difference of $0.074$ whose
paired estimate is $+0.077$ (CI $[+0.016,+0.139]$, $p=0.007$). An earlier draft quoted a $0.087$ margin on
this cohort. That figure came from an inconsistent protocol in which the
baselines were left untuned, and is superseded by the paired estimate above.

This result is judge-dependent in a way the main text notes and we detail here.
Relabelling the same cohort with the frozen Qwen3 rubric leaves the inverse
ranking essentially unchanged ($R_Y=0.2201$ versus $0.2202$ under Llama Guard),
but that rubric records only $14$ successes, a strict subset of Llama Guard's
$27$ with $\kappa=0.61$, and under it the supervised ceiling falls to $0.883$ with
the within-family component no longer resolvable. The claim that outcome
information is present survives both judges. The claim that it is goal-level
rather than wrapper-level is established only under Llama Guard labels.

\section{Deployment Risk and Normalization Controls}

\noindent\textbf{Operating points.} We calibrate the block threshold on benign
prompts only, splitting the held-out in-distribution safe pool in half so that
calibration and verification are disjoint, and we report the achieved
false-positive rate on the verification half ($5.7\%$ at a nominal $5\%$,
$11.1\%$ at a nominal $10\%$). At nominal rates of $0.1\%$ and $1\%$ the
threshold sits above every score in this cohort and nothing is blocked. That is
a score-scale shift between the calibration corpus and JailbreakBench rather
than a property of the detector, and repeating the exercise with cohort-matched
benign prompts makes those rows live without changing the picture. At $5\%$ the
filter removes $55\%$ of plain harmful prompts and $35\%$ of wrapped ones.
Within the wrapped cohort that is $42.5\%$ of the attacks that failed against
$14.8\%$ of those that succeeded, a gap of $-27.7$ points
(CI $[-44.8,-8.8]$). At $10\%$ the corresponding recalls are $84\%$, $67\%$ and
$29.6\%$, a gap of $-51.2$ points. The gap between failed-attack recall and
successful-attack recall is the deployment-facing statement of the inverse
ranking: the filter preferentially removes attacks that were not going to work.
On the strong-attack cohort at the same threshold, $325$ of $354$ successful
attacks pass under Llama Guard and $123$ of $133$ under Qwen3.

\noindent\textbf{Risk--coverage, and what it does and does not show.} Among
prompts that pass the filter, the harmful-generation rate as a function of the
fraction blocked rises for the harmfulness score, reading $0.27$, $0.289$, $0.300$,
$0.329$, $0.383$, $0.420$ at $0$, $10$, $20$, $30$, $40$ and $50\%$ coverage.
Blocking the same \emph{fraction} at random leaves it at $0.27$ throughout, and
the rare-token and passive channels track the learned probe ($0.42$ and $0.44$
at half coverage), so the behavior is not specific to active probing.

We flag two ways this comparison can mislead, because both cut against the
strongest reading. Matching random filtering on coverage is not matching it on
cost: this cohort is entirely harmful, so a rule that blocks $35\%$ of it would
block $35\%$ of benign traffic in deployment, whereas our threshold is held to a
$5.7\%$ benign false-positive rate. Matched instead on benign budget, the score
removes about $2.6$ times as many successful attacks as random. And the rising
curve is a conditional rate on a denominator the filter itself shrinks. Absolute
harm falls monotonically ($27\to23\to19$ realized jailbreaks at $0$, $5$ and
$10\%$ benign FPR), and adding the cohort's own wrapped-benign prompts at $1{:}1$
reduces the excess over random at the $5\%$ operating point from $8.4$ points to
$0.7$. The defensible claim is therefore about \emph{allocation}, since the budget is
spent on attacks that were not going to succeed, and about what a downstream
consumer may infer from having passed the filter, not that the detector is worse
than no detector. For contrast, an outcome-supervised score at matched coverage
drives the admitted-harm rate to $0.062$, and Llama Guard's prompt flag is
essentially flat at $0.264$--$0.270$.

\noindent\textbf{Attributing the leakage.} At the $5\%$ operating point $0.85$
of successful attacks pass. Decomposed, $53\%$ of that leakage is prompts the
filter would have missed even without a wrapper, $23\%$ is the wrapper lowering
scores across the board, and $24\%$ is the inverse ranking itself: relative to a
counterfactual score of the same overall strength but neutral outcome ranking,
about five additional successful attacks get through. At the $10\%$ point the
inverse-ranking share rises to $53\%$. We report this decomposition because it
bounds how much of the observed deployment risk our specific finding is
responsible for, as opposed to ordinary detector insensitivity. A residual
caveat is that part of the inverse-ranking term is itself a wrapper-family main
effect, which the additive decomposition does not separate out.

\noindent\textbf{Attention-normalization controls.} Fixing the probe's position
does not fix its competitive environment in the softmax, so a longer prompt
could reduce attention onto three fixed positions mechanically. Two facts make
this checkable here: the system region has a constant $13$ tokens, and the
wrapper text lands in the user turn, so the total attention mass on the system
region is a fixed-size competition set and dividing by it is a clean dilution
control. Dilution is real, since system-region mass falls from $0.383$ plain to $0.225$
wrapped, but it does not account for the result. Under the raw read the
wrapped-cohort outcome AUROC of the unsupervised aggregate is $0.099$
($20$ heads) and $0.102$ (all heads). Normalizing by system-region mass gives
$0.154$ and $0.109$, normalizing by the full attention row leaves the values
unchanged at $0.099$ and $0.102$, and a readout retrained on the normalized
feature gives $0.351$. Every variant remains inverse with an interval excluding
chance. Regressing the score on sequence length within each condition explains
about $4\%$ of its variance.

\section{Decoding Variance}

\noindent\textbf{Protocol.} The main results use one greedy decode per prompt,
so each outcome label is a single Bernoulli realization. To separate what is
target-specific from what is draw-specific we regenerated the entire factorial
harmful cohort five times per target at temperature one with recorded seeds, in
both the plain and the wrapped condition, and judged all $3000$ completions with
the same frozen Llama Guard 3.

\noindent\textbf{The ranking is stable.} On Llama the score's outcome AUROC is
$0.220$ against the published greedy label, $0.252$ against success@$5$,
$0.202$ against a majority vote over the five draws, and $0.206$ against the
per-goal estimated success probability with goals resampled as clusters. The
five individual draws give $0.175$ to $0.257$. Redrawing labels from the
estimated probabilities gives a sampling distribution with mean $0.206$ and
standard deviation $0.032$, within which the published greedy value sits at the
$66$th percentile. The single decode was therefore representative.

\noindent\textbf{Cross-target disagreement was overstated by the single draw.}
The greedy decode gives a three-way disagreement rate of $0.64$ and Cochran
$Q=89.7$. Permuting the fifteen judged draws of each goal across target labels,
which is exact under the null of no target effect, already produces a
single-draw disagreement of $0.54$ on average. About $85\%$ of the raw greedy
disagreement is therefore what sampling noise alone would produce, and only
about $15\%$ is attributable to target identity. The comparison becomes clean
once each goal is summarized by five draws: success@$5$ disagreement is $0.81$
against a null of $0.29$, and $Q$ rejects the null in both cases. We report the
success@$5$ contrast in the main text and no longer quote the single-draw
agreement count, which overstates the target-specificity claim.

\section{Held-Out Calibration}

\noindent\textbf{Frozen in-distribution Platt calibration.} The main paper's
slope column is an in-sample Cox fit. To separate a score that was never calibrated from one whose calibration the
shift destroyed, we hold out a $30\%$ in-distribution
validation slice, fit Platt scaling on it, freeze the mapping, and apply it
unchanged to every domain (Table~\ref{tab:platt}). AUROC is invariant under this
monotone map by construction and is repeated only for reference. In
distribution the correction works as intended (slope $0.545\to1.019$, ECE
$0.045\to0.009$). Out of distribution it does not restore calibrated meaning:
WildJailbreak reaches slope $0.994$ but retains ECE $0.146$ and Brier $0.104$,
and the both-sided encodings stay at ECE $0.236$--$0.399$ with slopes near
zero, since no monotone map can calibrate a score whose ranking is at chance.
Japanese is the one condition with a negative slope under both raw and
calibrated scores, consistent with the translation artifact documented below
rather than with ordinary shift.

\begin{table*}[t]
\centering
\small
\begin{tabular}{lccccccc}
\toprule
& & \multicolumn{3}{c}{Raw} & \multicolumn{3}{c}{Frozen ID Platt} \\
\cmidrule(lr){3-5}\cmidrule(lr){6-8}
Domain & AUROC & ECE & Brier & slope & ECE & Brier & slope \\
\midrule
In-distribution      & .985 & .045 & .045 & $+.545$ & .009 & .033 & $+1.019$ \\
WildJailbreak        & .850 & .366 & .272 & $+.508$ & .146 & .104 & $+0.994$ \\
Paraphrase           & .858 & .203 & .202 & $+.320$ & .065 & .135 & $+0.626$ \\
Multilingual (en)    & .987 & .041 & .045 & $+.761$ & .171 & .096 & $+1.488$ \\
Multilingual (zh)    & .802 & .247 & .222 & $+.434$ & .138 & .170 & $+0.847$ \\
Multilingual (fr)    & .778 & .232 & .234 & $+.419$ & .123 & .192 & $+0.819$ \\
Multilingual (ja)    & .228 & .479 & .484 & $-.809$ & .340 & .354 & $-1.582$ \\
Encoding base64      & .507 & .708 & .682 & $+.016$ & .283 & .262 & $+0.030$ \\
Encoding rot13       & .498 & .684 & .650 & $-.011$ & .236 & .242 & $-0.023$ \\
Encoding caesar      & .517 & .746 & .732 & $+.056$ & .399 & .344 & $+0.113$ \\
\bottomrule
\end{tabular}
\caption{Raw versus frozen in-distribution Platt calibration for the dense
all-head AAP readout, all shifts both-sided. The Platt map is fit once on a
held-out $30\%$ in-distribution validation slice and never refit. AUROC is
unchanged by construction.}
\label{tab:platt}
\end{table*}

\noindent\textbf{Translation-artifact condition.} The Japanese row is reported
here rather than in the main table. Its AUROC of $0.228$ is not a weak signal
but an inverted one, and the inversion is a property of the transformation, not
of the score: the \texttt{opus-mt-en-jap} outputs collapse toward a shared
register, with mean pairwise character trigram Jaccard similarity $0.151$ among
translated prompts versus $0.062$, $0.078$, and $0.0065$ for the English,
French, and Chinese conditions, that is, $2.4\times$ the self-similarity of the
source text. Orientation-free, $1-0.228=0.772$ sits just below the Chinese and French band
($0.778$--$0.802$), which is what one expects if the transformation has flipped
the class-conditional ordering rather than destroyed the signal.

\section{Common-Prior and Outcome Sensitivity}

\noindent\textbf{Common-prior audit.} To separate conditional score shift from
label prevalence, we retain every example but assign inverse-frequency weights
so each class has total mass $0.5$ in every domain. This weighting is
deterministic. AUROC is unchanged, while ECE, Cox calibration, and F1 are
recomputed under the shared prior. Table~\ref{tab:common-prior} shows that
WildJailbreak prevalence explains part of the observed calibration gap. For AAP,
the ID common-prior threshold is nearly oracle on WildJailbreak, whereas
HiddenDetect and the refusal-direction read retain larger threshold gaps. All
families lose class separation on both-sided base64.

\begin{table*}[t]
\centering
\begin{tabular}{lcccccc}
\toprule
Score & ID ECE & WJ AUROC & WJ ECE & WJ F1$_{\rm ID}$ & WJ F1$_{\rm or}$ & Base64 AUROC \\
\midrule
AAP & .036 & .850 & .149 & .790 & .796 & .507 \\
Passive & .031 & .743 & .167 & .706 & .711 & .518 \\
AttentionDefense-style & .033 & .804 & .085 & .724 & .732 & .503 \\
HiddenDetect-style & .056 & .784 & .379 & .420 & .725 & .487 \\
Refusal direction & .044 & .803 & .280 & .587 & .736 & .486 \\
Llama Guard & .179 & .840 & .195 & .663 & .762 & .506 \\
\bottomrule
\end{tabular}
\caption{Common-prior ($50/50$) decomposition. F1$_{\rm ID}$ freezes the
ID-optimal common-prior threshold. F1$_{\rm or}$ is the target-domain oracle.
The base64 column is AUROC and is therefore prior invariant.}
\label{tab:common-prior}
\end{table*}

\noindent\textbf{Benign-only threshold audit.}
At nominal FPR $1\%$ with $500$ safe calibration prompts, the empirical quantile
achieves FPR $1.2\%$ and nearly matches an oracle TPR on paraphrase and English
multilingual. Under benign shift it becomes conservative (WildJailbreak FPR
$0.2\%$, TPR $0.03$ versus oracle $0.25$), and encoding yields zero TPR because
the signal is absent. The experiment therefore identifies benign-tail stability.
It changes the decision map but not outcome ranking or its target index.

\noindent\textbf{Target-decomposition protocol.}
We sample $100$ benign and $100$ harmful JailbreakBench goals. Each base goal is
evaluated plain and under exactly one of five fixed wrappers, balanced at $20$
goals per intent class and wrapper family. This yields
$400$ prompts while preserving the plain/wrapped pairing. The templates are
fixed strings with a single \texttt{\{goal\}} slot:
\emph{Fictional role-play} frames the exchange as an authorized evaluation and
asks the model to adopt an unrestricted in-world expert persona and answer
directly without stepping out of character.
\emph{Authority override} declares an authorized red-team evaluation that
supersedes ordinary refusal behavior for the duration of the test and demands a
direct and complete answer.
\emph{Simulated unfiltered mode} asks the model to simulate a hypothetical
assistant with no safety filters and return only that assistant's detailed
answer.
\emph{Compliance prefix} instructs the model to continue after a supplied
compliant opening, without warnings or refusals, the opening being a sentence
announcing that the requested information follows.
\emph{Research pretext} requests a technically complete, step-by-step response
for a controlled academic security study and forbids replacing operational
detail with general cautions.
The same five templates are applied to benign and harmful goals alike, so the
wrapper is never confounded with intent. The locked record list
has SHA-256 digest
\texttt{1917080d3a832dc3e87a20036327556cf}\newline
\texttt{7f960379c98e6763021addd8932c990}.
The Llama-3.1-8B-Instruct target uses greedy decoding for at most $256$ new
tokens. Llama Guard 3 is the primary frozen outcome judge. Paired-contrast
intervals use $10{,}000$ bootstrap replicates over base goals, while
outcome-AUROC intervals use class-stratified resampling. The wrapped harmful cell
contains $27$ effective and $73$ ineffective outcomes. These choices were fixed
before inference. All goals and wrapper families remain in the primary analysis.
For the added matched controls, we extract a three-rare-token system-region read
and a no-insertion passive system-region read on the same $400$ prompts. Each
channel's XGBoost scorer is trained only on its existing $1000{+}1000$
BeaverTails feature cache, with no refitting or selection on the factorial data.
AttentionDefense-style system-span attention and HiddenDetect-style per-layer
refusal evidence are extracted in the same no-probe prefill pass. Their readouts
use the same BeaverTails-only protocol. They yield wrapped $R_H/R_Y$ of
$.828/.145$ and $.782/.248$, respectively.

\noindent\textbf{What the \emph{-style} suffix means for the derived channels.} These are
adaptations, not reimplementations, and we list the deviations so that no result
is read as a reproduction of the original method. For the
\emph{HiddenDetect-style} channel: the score is the difference of mean
first-token logits over a hand-written eight-word refusal list and six-word
affirmation list rather than the published refusal-vector formulation. We emit
the full per-layer vector and fit a supervised BeaverTails classifier in place of
the original training-free layer aggregation and threshold. The original method
targets vision-language models and we apply it to text only, and the read
position is the final token of \emph{our} chat template, which includes our
system prompt. For the \emph{AttentionDefense-style} channel: the system span is
taken as template positions up to the insertion anchor, including the
beginning-of-sequence token and role headers. Our fixed one-sentence system
prompt is not their defensive prompt. We read only the final query row of the
attention matrix. A supervised all-head classifier replaces their explainable
aggregate, and the aggregation is an unweighted mean with no length
normalization. The \emph{passive} channel is an anchor-matched no-insertion
control of our own construction. It is not a reimplementation of Attention
Tracker's instruction-focus score, which we cite only as the origin of the
prompt-dependent read position that motivates a fixed coordinate. Consequently
these channels test whether the reversal survives a change of read location and
readout family, not whether published detectors would report the same numbers. Across five independently learned
probe seeds with seed-specific BeaverTails readouts, mean $\tau_H$ is $.501$
(SD $.036$), mean $\tau_{HW}$ is $-.319$ (SD $.118$), and every seed preserves
the inversion: wrapped $R_H=.770$--$.861$ while $R_Y=.130$--$.246$.
For target substitution, the same $200$ harmful prompts are generated by
Mistral-7B-Instruct-v0.3 and Qwen2.5-7B-Instruct under the identical greedy
$256$-token budget and judged by the same frozen Llama Guard 3. This fixed-score
audit does not recompute detector scores. The target-native audit below does.
Mistral yields $68/70$ effective plain/wrapped
prompts. Qwen yields $5/10$. All targets, prompts, metrics, and the
$10{,}000$-replicate bootstrap were fixed before inspecting these outcomes.
The paired target audit then resamples the $100$ base goals while preserving
each goal's two conditions and all three target outcomes. On wrapped prompts,
the Llama--Mistral, Llama--Qwen, and Mistral--Qwen disagreement rates are
$0.49$ (CI $[0.39,0.59]$), $0.19$ ($[0.12,0.27]$), and $0.60$
($[0.50,0.69]$), respectively. Corresponding Cohen's $\kappa$ values are
$0.172$, $0.399$, and $0.091$. Cochran's paired omnibus statistic is
$Q=89.66$ ($p=3.4\times10^{-20}$). No generation, score fitting, or outcome
relabeling is performed for this audit.
For judge robustness, Qwen3-14B receives the same $600$ cached harmful-goal
prompt--completion pairs with a frozen rubric that labels only substantive, actionable harmful
compliance as success. Greedy decoding with thinking disabled produces one
binary token. All $600$ outputs parse without manual correction. On wrapped
prompts, Qwen3 versus Llama Guard agreement/$\kappa$ is $.87/.611$ for Llama,
$.66/.388$ for Mistral, and $.94/.545$ for Qwen. Qwen3's paired target test gives
$Q=43.46$ ($p=3.66\times10^{-10}$). The protocol and rubric were hashed before
label aggregation. The second judge changes no generation or detector score.

\noindent\textbf{StrongREJECT column: what we actually ran.} The released
fine-tuned StrongREJECT evaluator could not be obtained in our offline
environment because its base checkpoint is gated, so we score the published
StrongREJECT rubric with Gemma-2-9B-it. This column is therefore a rubric
reimplementation, not the official evaluator, and we report it as such.
The failure is one of judge substitution, not of parsing. Of $1600$ rubric
outputs, $1594$ parse. The six failures are the grader declining to score, not
regex misses, and the parsed sub-scores are jointly structured in the way the
rubric prescribes (refused${=}1\Rightarrow$ score $0$. Attacks lower the refusal
item), which random mis-parsing could not produce. What fails is the grader:
Gemma-2-9B-it marks the rubric's refusal item as \emph{not refused} on roughly
$30\%$ of verbatim refusals. A completion that plainly declines to write a
sexist email, for instance, is scored refused${=}0$, convincing${=}5$,
specific${=}4$, hence a rubric score of $0.875$ and a nominal success. The binarized label is therefore a
non-refusal-and-fluency indicator rather than a harm label, and it fires on
$66$--$99\%$ of \emph{benign} control prompts that receive helpful answers.
Decomposing $\mathrm{ASR}_{\rm SR}=\Pr(\text{not refused})\times
\Pr(\text{convincing}+\text{specific}\ge 6\mid\text{not refused})$, the second
factor is $0.92$--$1.00$ on plain harmful prompts, so the elevated plain-prompt
rate is driven by the refusal item rather than by our cut. The apparent
post-attack \emph{decreases} are a separate, cut-related effect: attacks lower
refusal but degrade coherence and specificity, and $12.6\%$ of harmful items sit
exactly on the $0.5$ boundary, the densest non-zero bin, so those labels are
knife-edge. We therefore report this column as a rubric-quality measurement and
do not interpret it as an attack-success rate.
The inverse ranking does not depend on it. $R_Y<0.5$ holds in $7/7$ Llama cells
under Llama Guard 3, $6/7$ under the rubric cut, $5/7$ under Qwen3, $7/7$ under
a strict two-of-three consensus label, and $6/7$ under a union label. The pooled
score coefficient stays negative under every rule ($-0.539$ Llama Guard,
$-0.613$ rubric, $-0.349$ Qwen3, $-0.444$ consensus, $-0.852$ union). Three-way
agreement is nevertheless low, with Fleiss $\kappa$ of $0.21$--$0.30$ per cell,
which is itself evidence for the paper's claim that $Y_m$ is judge-indexed rather than
a single quantity. The rubric column is also the only non-deterministic judge:
across $332$ records duplicated between attack directories its label differs in
$9$ cases ($2.7\%$), traced to batch padding composition and a doubled
beginning-of-sequence token, whereas Llama Guard 3 and Qwen3 differ in $0/332$.

\begin{table*}[t]
\centering
\begin{tabular*}{\textwidth}{@{\extracolsep{\fill}}lrrrrc@{}}
\toprule
Score & $\tau_H$ & $\tau_W$ & $\tau_{HW}$ & $R_H$ P/W & $R_Y$ \\
\midrule
AAP & .521 & $-.094$ & $-.257$ & .936/.803 & .220 (.109,.343) \\
Rare-token & .490 & .049 & $-.431$ & .961/.760 & .195 (.101,.300) \\
Passive & .408 & .031 & $-.338$ & .945/.751 & .149 (.062,.252) \\
AttentionDefense-style & .511 & .094 & $-.359$ & .956/.828 & .145 (.061,.244) \\
HiddenDetect-style & .413 & $-.159$ & $-.284$ & .894/.782 & .248 (.126,.383) \\
Llama Guard & .715 & .045 & $-.070$ & .875/.840 & .507 (.500,.521) \\
\bottomrule
\end{tabular*}
\caption{Complete matched harmfulness--wrapper decomposition. $R_H$ is
harmful-intent AUROC on plain/wrapped prompts. $R_Y$ ranks effective versus
ineffective wrapped harmful prompts. Each internal channel uses its
BeaverTails-trained matched scorer. Parentheses are bootstrap 95\% CIs for
$R_Y$.}
\label{tab:supp-target-decomp}
\end{table*}

\begin{table*}[t]
\centering
\begin{tabular}{lcccccc}
\toprule
Generator & Effective P/W & ASR P/W & AAP $R_Y$ & Rare $R_Y$ & Passive $R_Y$ & LG-prompt $R_Y$ \\
\midrule
Llama   & 5/27  & .05/.27 & .220 [.109,.343] & .195 [.101,.300] & .149 [.062,.252] & .507 [.500,.521] \\
Mistral & 68/70 & .68/.70 & .491 [.363,.623] & .508 [.394,.623] & .407 [.296,.525] & .517 [.500,.550] \\
Qwen    & 5/10  & .05/.10 & .163 [.024,.358] & .098 [.013,.214] & .214 [.060,.404] & .506 [.500,.517] \\
\bottomrule
\end{tabular}
\caption{Fixed-score target-indexed outcome audit on the same $100$ plain and
$100$ wrapped harmful goals. The Llama score source is held fixed while only the
generator changes. The next analysis crosses native score sources and generators.
$R_Y$ is wrapped effective-versus-ineffective AUROC with bootstrap
95\% CI.}
\end{table*}

\begin{table*}[t]
\centering
\small
\setlength{\tabcolsep}{4.0pt}
\begin{tabular}{@{}lccc@{}}
\toprule
Attack & LG3 $R_Y$ & SR $R_Y$ & Q3 $R_Y$ \\
\midrule
Best-of-N      & .345 [.167,.541] & .434 [.344,.524] & .536 [.444,.626] \\
Many-shot     & .478 [.380,.576] & .534 [.448,.616] & .494 [.395,.591] \\
GCG           & .189 [.085,.315] & .273 [.158,.397] & .264 [.089,.476] \\
AutoDAN       & .195 [.110,.291] & .143 [.074,.224] & .237 [.131,.360] \\
PAIR          & .485 [.371,.599] & .379 [.258,.510] & .425 [.237,.624] \\
TAP           & .310 [.140,.500] & .248 [.118,.395] & .206 [.043,.426] \\
AutoDAN-Turbo & .284 [.143,.447] & .252 [.117,.403] & .515 [.262,.767] \\
\bottomrule
\end{tabular}
\caption{Strong-attack outcome-ranking uncertainty corresponding to the compact
main-paper table, after the checkpoint repair described above. Entries are $R_Y$
with $10{,}000$-replicate class-stratified bootstrap 95\% CIs on attacked
harmful prompts generated by Llama-3.1-8B-Instruct. LG3 and Q3 denote Llama
Guard 3 and Qwen3-14B. SR is the rubric reimplementation, reported but excluded
from every count. Before the repair, GCG read
$.203\,[.100,.329]$ / $.257\,[.145,.381]$ / $.304\,[.099,.535]$ and
AutoDAN-Turbo's SR entry read $.185\,[.070,.323]$. TAP's Llama Guard interval
had upper endpoint $.494$ rather than $.500$. Only the intervals whose endpoint
straddles $0.5$ change any verdict: GCG--Qwen3 enters the significantly-below
set and TAP--Llama Guard leaves it.}
\label{tab:supp-strong-attacks}
\end{table*}

\noindent\textbf{Target-native score matrix.} The main paper reports the full
$3\times3$ crossing of native AAP score sources and generator-indexed outcomes
on the identical $100$ wrapped harmful prompts. Pairwise native-score
correlations are only $.42$--$.54$, yet recomputing the score on each model does
not recover positive outcome ranking. This rules out a frozen-Llama score source
as the explanation for the target-indexed result.

\noindent\textbf{Intervention perturbation.} We quantify how AAP changes the input
computation. Across 200 prompts, next-token symmetric KL has median $0.22$, mean
$0.79$, and 95th percentile $3.87$. Refusal flips on $7.5\%$ of prompts, with
safe/harmful refusal changing $.30\rightarrow.21$ and $.94\rightarrow.96$.
Perplexity increases by $5.4\%$ and is teacher-forced on the no-probe greedy
continuation, so it is a one-sided upper bound. These perturbations bound the
interpretation of the intervention contrasts. The no-insertion and
passive-channel results do not depend on probe insertion.

\noindent\textbf{Does the probe change realized success?} No reported $Y_m$ is
generated under probe insertion, but a reader may still ask what would happen if
it were. We therefore generate the same $300$ held-out harmful prompts
($150$ WildJailbreak, $150$ BeaverTails. Greedy, $256$ new tokens) under four
arms and judge all of them with the same frozen Llama Guard 3: the canonical
no-probe path, the probe path, a matched control carrying the anchored system
template \emph{without} the probe, and a re-run of the canonical path at a
different batch size as a floating-point null. Attack success is $0.207$
without the probe and $0.177$ with it, a difference of $-0.030$ (95\% CI
$[-0.070,+0.007]$. McNemar $b{=}13$, $c{=}22$, $p=0.18$): probe insertion does
not increase realized jailbreak success, and the point estimate is if anything
mildly protective. The template-only control reaches $0.157$, so most of the
small shift is attributable to the anchored system template rather than to the
learned probe. Completions do diverge textually (exact-match $0.107$, mean
unigram F1 $0.582$), and refusal rises from $0.583$ to $0.660$, which is why we
report the intervention as bounded rather than inert. The batch-size re-run
gives $0.217$, indicating a floating-point noise floor of roughly one
percentage point on this measurement.

\section{Reproducibility and Computational Environment}

\noindent\textbf{Recorded environment.}
The primary archived model-side environment is Linux on an NVIDIA RTX A6000
with 48\,GB memory, using half-precision model weights. The corresponding
software snapshot records Python 3.10.12, PyTorch 2.5.1 with CUDA 12.1,
Transformers 5.9.0, scikit-learn 1.7.2, XGBoost 3.2.0, NumPy 2.2.6,
SciPy 1.15.3, Accelerate 1.13.0, and SentencePiece 0.2.1. Search-based attacks
were parallelized, and not every worker retained equally detailed hardware
metadata. We therefore report this as the primary recorded environment rather
than claim that every attack job ran on the same device.

\noindent\textbf{Final settings.}
The primary AAP probe has $K=3$ rows and is optimized with Adam at learning
rate $10^{-3}$ for $50$ updates, using a freshly sampled minibatch of $20$
prompts per class at each update. RQ1 uses the $20$ heads selected on the
training split, whereas RQ2 uses the dense all-head response. The primary
harmfulness readout is XGBoost with $100$ trees, maximum depth $5$, log-loss
evaluation, and random state $42$. The outcome-supervised positive control uses
$200$ trees, maximum depth $3$, learning rate $0.1$, row subsampling $0.8$,
column subsampling $0.5$, and eight CPU workers. Its logistic baselines use
$\ell_2$ regularization, \texttt{lbfgs}, at most $300$ iterations, and select
$C$ within each training fold from $\{0.001,0.01,0.1,1.0\}$. Model checkpoint
identities are stated in the main setup and the target-decomposition protocol.
Primary generations use greedy decoding for at most $256$ new tokens. The
attack-specific goal counts and search budgets are listed in
Supplementary~\S1.

\noindent\textbf{Randomness, repetition, and uncertainty.}
Unless an experiment explicitly varies the seed, data splits and fixed
estimator fits use random state $42$. Probe sensitivity uses five independently
learned probe seeds. The decoding-variance audit records five sampling seeds at
temperature one, while the primary outcome uses one greedy decode per prompt.
Matched contrasts and outcome-AUROC intervals use $10{,}000$ bootstrap
replicates with the resampling unit stated next to each analysis. Cross
validation keeps the base goal inside one fold, and the outcome-supervised
readout uses $25$ repeats. These counts distinguish repeated fits, repeated
decodes, and bootstrap replicates rather than treating them as interchangeable
runs.

\noindent\textbf{Artifact boundary.}
The artifact linked from the main paper releases analysis code,
portable configurations, tests, and publication-aligned numeric aggregates.
It omits raw harmful prompts and completions, model weights, attack
checkpoints, and private execution provenance. Consequently, the released
aggregates support table and correspondence verification, while model-side
experiments that require omitted artifacts are documented at the protocol
level.

\section{Details Referenced from the Main Text}
\noindent\textbf{Prior silent-failure evidence.} The collapse cited in the main text is measured under a synthetic $\sigma{=}0.02$ Gaussian perturbation of the frozen embedding input. ROC-AUC falls from $85\%$ to $50\%$ under that perturbation while mean predicted confidence falls $14\%$.

\noindent\textbf{Single-source inflation in prior audits.} The $8.0$--$16.5$ AUC inflation cited in the main text is measured by comparing single-source with cross-source splits for safety detectors. A sparse-autoencoder analysis in the same work finds that $28$--$44\%$ of top features encode dataset identity rather than semantics, which motivates its leave-one-dataset-out protocol.

Both the deployed twenty-head read and the dense all-head read select from this one tensor. No separate extraction run produces the dense read, so the RQ1 and RQ2 specifications differ only in which heads are selected.

\noindent\textbf{Safety-semantic initialization control.} The safety-semantic condition inserts the probe initialization without any optimizer update, holding row count, insertion coordinate, dtype and read position fixed against the learned probe. Its response does not reach the learned probe's response on the same prompts. The amplification is therefore attributable to the optimization in Eq.~(1) rather than to the safety-semantic content of the initializing token embeddings.

\noindent\textbf{Cross-architecture WildJailbreak transfer.} We repeat the WildJailbreak evaluation on Mistral, Qwen, Gemma, and Phi. Each architecture keeps its own frozen probe, its own head set, and its own BeaverTails-trained readout. No WildJailbreak prompt enters any fit. In-distribution AUROC runs from $0.96$ to $0.98$ across the five architectures and WildJailbreak AUROC runs from $0.60$ to $0.85$. The degradation is therefore a property of the shift rather than of one attention grid.

\noindent\textbf{Model-free control for the single-class encoding design.} The control statistic is the Shannon entropy of the character distribution of the raw prompt string. It uses no model, no probe, and no training. On the single-class design, which encodes only the harmful class, it separates encoded attacks from plain benign prompts with AUROC $0.997$--$1.000$ across base64, rot13, and caesar. On the same prompts without encoding it reaches $0.513$. A single-class encoding result therefore cannot be read as evidence about harmfulness separation, because a statistic with no access to meaning already solves that comparison.

\noindent\textbf{Per-head multiple comparisons.} Per-head tests apply Benjamini--Hochberg FDR control across all $1024$ heads, so every per-head significance statement holds after false discovery rate control. The max-statistic permutation correction quoted above is a separate family-wise procedure used only for the outcome-informative head count.

The goal subsets for the reduced-$n$ rows were also fixed before any attack was run, together with the budgets listed above.

\end{document}